# DeepCloud

**Ardavan Bidgoli**
Carnegie Mellon University

**Pedro Veloso**
Carnegie Mellon University

The application of a data-driven, generative model in design

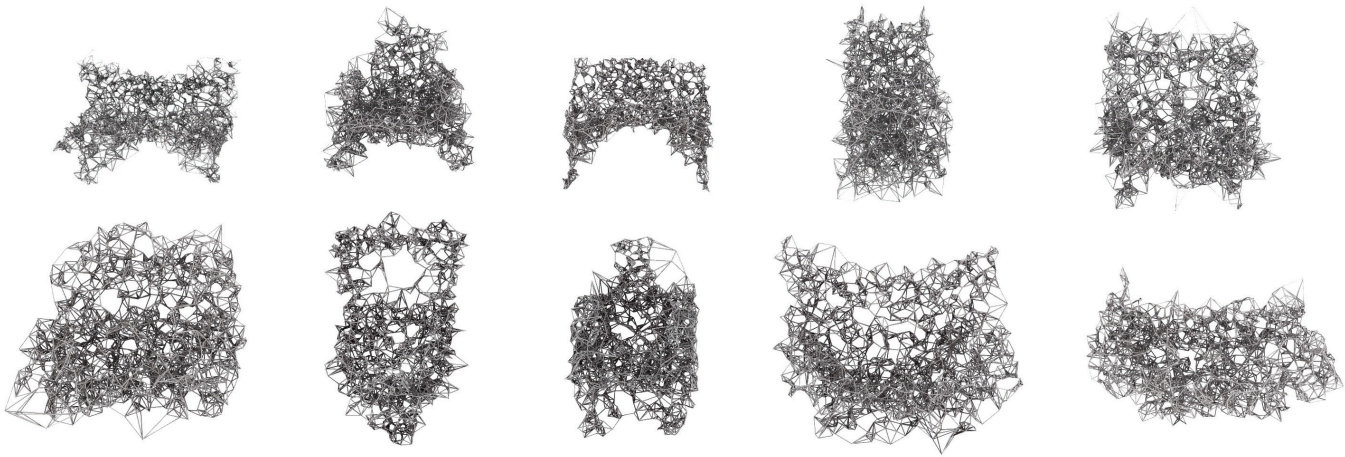

1  Data-driven, generated chair samples

## ABSTRACT

Generative systems have a significant potential to synthesize innovative design alternatives. Still, most of the common systems that have been adopted in design require the designer to explicitly define the specifications of the procedures and, in some cases, the design space. In contrast, a generative system could potentially learn both aspects through processing a database of existing solutions, without the supervision of the designer. To explore this possibility, we review recent advancements of generative models in Machine Learning and current applications of learning techniques in design. Then, we describe the development of a data-driven generative system titled DeepCloud. It combines an autoencoder architecture for point clouds with a web-based interface and analog input devices to provide an intuitive experience for data-driven generation of design alternatives. We delineate the implementation of two prototypes of DeepCloud, their contributions, and potentials for generative design.

## INTRODUCTION

In a conventional design process, the designer should explicitly address the design problem and explore solutions (3 top). On the other hand, designers can also develop an automated system to generate design alternatives (3 middle). Such systems are denominated Generative Systems (Mitchell 1977; Fischer and Herr 2001). Originally, generative systems for design incorporated Artificial Intelligence problem-solving procedures, such as search, optimization and linear programming, or syntactic formulations, such as shape grammars (Mitchell 1977, 425–474). In the past decades, the repertoire of generative

|  | Fractal Geometry | L-System | Shape-grammar | Cellular Automaton | Search | Optimization | Agent-based model | Physics simulation |
|---|---|---|---|---|---|---|---|---|
| Procedure | Application of rules to substitute shape | Application of rules to substitute character in string and mechanism to convert string to shape. | Application of rules to derive a shape | Change of state for cells in time-steps, parametrized by the state of the neighborhood | Strategies to select actions for discrete states of the solution. | Strategies to navigate in the design space and look for solutions with better fitness score. | Simulation of agents that follow local rules in environment along time-steps. | Simulation of physical forces in geometrical bodies in discrete time-steps |
| Design space | State space: set of all shapes reachable from the initial shape by the available rules | State space: set of all strings reachable from the axiom by the production rules. | State space: set of all shapes reachable from the initial shape by the available rules | State space: all joint states of the cells reachable from the initial joint state by the available rules. | State space: The set of all states reachable from the initial state by the available actions | State-space landscape: all valid solutions parametrized with a score for an objective function. | Consequence of interaction | Consequence of interaction |

procedures has been primarily extended through a diverse set of computational techniques, some of which were evolutionary heuristics for optimization, agent-based models, and physics simulation (2).

The use of these generative procedures implicitly or explicitly configures the scope of the possible design alternatives, also known as design space (2). For example, in search algorithms, the design space is a graph called state space, which represents "the set of all states reachable from the initial state by any given sequence of actions" (Russel and Norvig 2010, 67). In the optimization approach, it is a state-space landscape that represents all the parameters of the solution as a location in the landscape, and the value of the solution in respect to the adopted metric as the elevation (p. 121).

Despite the potential of these systems to generate design alternatives, in their canonical form, the designer needs to specify their procedures. Namely, the designer composes a generative system based on an interpretation of the problem and on the choice of strategies to generate solutions. In some cases, such as optimization, even the design space is specified via parametrization in a modeling application, providing "all the unique formal possibilities of a given design model" (Nagy 2017).

Nevertheless, an intelligent system could potentially learn both the design space and the procedures to navigate it with previous experiences or solutions to a given problem and without the supervision of the designer (3 bottom).

Data-driven learning is a core topic of Machine Learning (ML), a multidisciplinary field concerned with the question of "how to construct computer programs that automatically improve with experience" (Mitchell 1997, xv). To solve many of the learning tasks, researchers employ discriminative and generative models. After the training, a discriminative model only learns how to solve the learning task (namely; classification, regression, clustering, dimensionality reduction, etc.). Given a new input, it only provides the respective

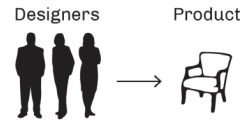
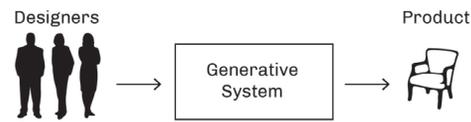
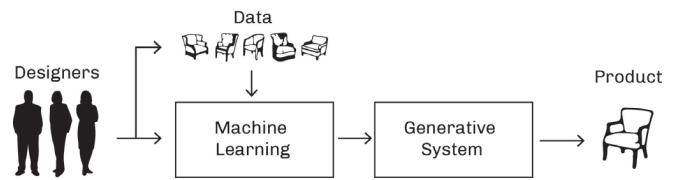

2  Three common generative systems in design

3  Top: Traditional design process; Middle: Generative design process; Bottom: Data-driven generative design process (based on Fischer and Herr 2001, 3).

output. In probabilistic terms, it strictly learns the posterior probabilities: the probability of the output, given the input. In contrast, after the training phase, a generative model "explicitly or implicitly model[s] the distribution of inputs as well as outputs" (Bishop 2006, 43). More than learning how to perform the required task, it models how the data has been generated. Thus, a generative model enables the sampling of synthesized data based on the data distribution that it learned for the task – i.e., it creates a data-driven generative system.

Despite the potential of the data-driven generative systems, there is almost no design application based on databases and few researchers have investigated design exploration of geometric models with data-driven, generative systems. Due to this gap, there are no standards for modes of interaction, performance evaluation, and design representation with ML models for generative design. This paper addresses this gap by prototyping a design tool based on a data-driven, generative model from ML, with which the designer can interact and generate new forms.

## LITERATURE REVIEW

To understand this gap, we will review some of the generative models in ML that can contribute to design synthesis. Then, we will provide a brief review of the recent application of ML techniques in generative systems.

### Generative models in Machine Learning

There are several generative models in ML, including but not limited to: Principal Component Analysis, Autoencoder, Variational Autoencoder, and Generative Adversarial Networks. In this section, we briefly describe some of these models.

Principal Component Analysis (PCA) is used for tasks such as dimensionality reduction by orthogonally projecting the data of a high dimensional space onto a lower dimensional linear space, called hyperplane. Sampling the points on this hyperplane can potentially generate new samples with similar characteristics as the input.

Another generative model used for dimensionality reduction is the Autoencoder (AE). It is an artificial network composed of two parts: an encoder and a decoder. The encoder learns how to compress the samples of a data distribution into a smaller latent representation while preserving its structure. The decoder learns how to reconstruct the original input just by observing this latent representation and, later, can be used to synthesize new output data. The advantage, compared to the PCA method, is that the autoencoder can learn to preserve the nonlinear structure of the input data.

Variational Autoencoders (VAE) are resembling AE structure, but with a substantially different mathematical back-end. In VAE, the encoder learns a probability distribution (a latent variable model) rather than a random compressing function (Kingma and Welling 2013). The decoder samples from this probability distribution and learns to reconstruct the original input.

Generative Adversarial Networks (GAN) (Goodfellow et al. 2014) are generative models created with the primary purpose of synthesizing new data that fit into a probability distribution. GAN architecture leverages two adversarial neural networks, a generator, and a discriminator. The discriminator is trained with the samples from a dataset and learns to determine if a given input belongs to it or not. In contrast, the generator has no access to the dataset.

As a result, the generator masters the task of synthesizing data that fits in the distribution of the database. GANs gained significant attention not only in the field of AI but also in the arts. By introducing variations to GAN architecture, different synthesis procedures can be achieved[1].

### Machine Learning in generative design

While ML provides techniques that can be directly applied to the quantitative analysis of buildings, recently they have also been incorporated as a component of generative systems.

Data-driven models for dimensionality reduction have been used as tools to embed the generated solutions of an optimization procedure in a lower dimensional map. For example, Koenig, Standfest, and Schmitt (2014) used a self-organizing map (SOM) to reduce the dimensions of the solutions generated by a multi-criteria optimization of building blocks to a two-dimensional grid.

The same type of visualization, mostly based on statistical methods, is presented in the project DreamLens, by Autodesk (Matejka et al. 2018), to enhance users' ability to navigate in a high-dimensional design space generated by project DreamCatcher (Nourbakhsh 2016). It helps the user interactively map thousands of solutions to a two-dimensional space for better navigation of a design space.

However, these applications of ML are restricted to visualization and do not solve any task in the generative system. On the other hand, Sjoberg, Beorkrem, and Ellinger (2017) adopted ML techniques both to visualize and to support design optimization. Their workflow incorporates a supervised neural network to predict the user selection of the input population for a Genetic Algorithm (GA). The neural network becomes the fitness function used by the GA to produce the next generations. Additionally, a PCA is combined with a density-based spatial clustering for visualization clusters with high performance and their respective average solution in three-dimensional space.

Harding and Derix (2011) developed a system to generate the layout of an exhibition hall for multiple exhibitions that contains two ML components. First, a SOM embeds the multi-dimensional feature space of the exhibited objects of each future exhibition in a separate two-dimensional map. They connect the closest neighbors of the objects of the grid, creating a planar graph for each exhibition. The second component, a growing neural network, clusters the multiple graphs of the future exhibitions according to similar topologies using their spectrum for the Laplacian matrix.

Zaghloul (2015) used an SOM as a method to explicitly generate and organize new design alternatives of a villa. Zaghloul encodes the different spatial units of the villa as boxes that can be added or subtracted from the whole volume. The initial input of the SOM encodes the six design alternatives for the villa. The output layer of the SOM is a two-dimensional grid of 15 by 15 cells containing the original input and generated non-linear morphing samples between them.

Narahara (2017) developed a multi-agent adversarial learning experiment that builds urban blocks in a three-dimensional grid. Ten competing color-coded groups of agents use a shallow neural network with custom weights to define their actions. The inputs of the network are the four features detected in the cone of vision (ground level, teammate, opponent and/or building block) while the output is the probability of executing the four actions (step, flock, attack or build). The experiment is repeated multiple times. After each episode, the set of weights of the neural networks of the best four teams (teams that built more blocks) is preserved and they are recombined and mutated for the remaining teams in the next episode.

## OBJECTIVES

Two of the works presented in the previous section use ML technique as the main component of a generative system: Narahara (2017) uses a reinforcement learning approach to learn policies for an agent-based system and Zaghloul (2015) uses a SOM to create a map that generates non-linear morphing of the geometric input.

In this paper, we will address a generative system similar to the latter, focusing on the modeling of geometric solutions. In the following sections, we investigate a data-driven system that can generate new geometric models for design, concentrating on the following contributions:

- Data-driven design space: explore techniques that can learn the design space, not from parameterizations made by the designer, but from a potentially large database of examples.
- ML techniques: explore and employ the recent advancements of deep generative models in ML.
- Representation methods: use of a generic and flexible system of geometric representation that can address modeling geometric solutions for multiple problem domains.
- Operative space: instead of focusing on the visualization in lower-dimensional space, exploring the capacity of higher dimensional space to embed meaningful geometric transformations.

## METHOD

This section contains the research decisions that supported the development of our data-driven generative system.

After reviewing deep neural networks literature in search of potential architectures that could learn with point cloud data, we opted to use the autoencoder (AE) developed by Achlioptas et al. (2017), which achieves accurate reconstruction. In the first section, we will explain their approach to design the core ML model that we implemented as the back-end of DeepCloud with some modifications. Afterwards, we will focus on the architecture of DeepCloud and the front-end.

### The back-end:

For geometric representation, Achlioptas et al. opted for a representation of 3d forms that can be generalized to different problem domains: point clouds. In contrast to other methods of 3d representation like meshes, or 2d representation like multi-view images, point cloud has several advantages: 1) it is a compact, expressive, and homogeneous 3d representation 2) it is flexible to geometric operation, and 3) it can be produced either by sampling from existing digital models or by scanning physical objects using off-the-shelf 3d devices- i.e., Kinect and LIDARS scanners (ibid).

For the ML model architecture they opted to use an AE, due to the overall simplicity, ease of training, and the possibility of manipulating/controlling the outcomes. In contrast, RNNs (Pascanu 2013) and GANs are expensive computationally and hard to train (Achlioptas et al. 2017). Moreover, in the case of GANs, there is no possibility to establish a meaningful control scheme over the outputs.

Point cloud representations are still a challenging topic for AEs. The lack of an underlying ordering structure limits the application of convolution operators, which are very efficient in fixed representations, such as images. Besides, the reconstruction of a point cloud is not trivial, since there is no universal evaluation function to compare two sets of points. The architecture and metrics proposed by Achlioptas et al. (2017) resulted in a state-of-the-art reconstruction quality and generalization ability for 3d point clouds.

This AE learns how to compress a point cloud of 2048 points (2048 x 3 coordinates) into latent vectors of size 128 in the bottleneck layer. The architecture of the encoder is agnostic to permutations of the point clouds (4). A series of blocks with 1D convolutions with increasing number of

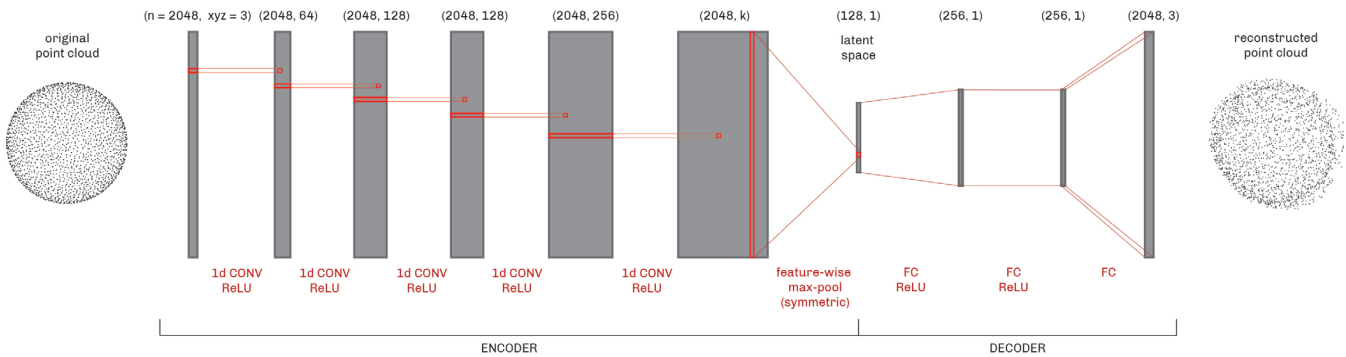

4   Architecture of the Autoencoder, based on (Achlioptas et al. 2017)

features combined with linear rectified linear units (ReLu) learns filters that are activated by the coordinates of each point of the input point cloud individually. To generate a joint-representation, a max-pool layer assign the maximum value of each $i^{th}$ column of the last convolutional block to the $i^{th}$ position of the latent vector in the bottleneck.

The decoder learns to reconstruct the initial point cloud from the latent vector. It has a very simple architecture composed of two fully connected layers combined with ReLu and a final fully connected layer that reconstructs the point cloud.

In order to train the AE, the authors adopted two different permutation-invariant metrics: Earth Mover's Distance and Chamfer Distance. The former is a bijection that measures the minimum total distance required to transform a point cloud into another. The later measures the sum of the squared distance between each point in one point cloud to the nearest neighbor of the other point cloud.

### DeepCloud/front-end

In addition to the implementation of the AE, the DeepCloud project adds a GUI with intuitive tools for the manipulation of a high-dimensional, latent space. Therefore, the learned latent space is not only the design space but also a modeling tool with multiple operations to transform the point cloud. It can potentially trigger the designer's imagination by supporting the generation of novel design objects.

## DEVELOPMENT
### Prototype 1

The first prototype of the DeepCloud was developed using a server-client application of Python/Tensorflow as the back-end and Rhino/Grasshopper as the front-end[2] (6). To provide a more intuitive experience, we opted in for a Leap Motion sensor as the main user input device.

The first step in DeepCloud is to set up and train the AE. In our setting, the encoder should learn how to compress a cloud of 2048 points into vectors of size 32 (latent space) and the decoder should be able to reconstruct the initial point cloud using the Chamfer Distance as the accuracy evaluation metric. In $R^{32}$, point clouds with similar characteristics are represented with similar latent vectors.

As a generative model, it can reconstruct the original objects from the database in addition to generating new synthesized objects. Navigating over each parameter of the latent space vector, the resulted point cloud demonstrates specific behavior on one or multiple features.

Each dimension of latent vector is associated with certain characteristics. For example, if the model has been trained on a dataset of chairs, it is possible to observe one element of the latent vector picks a feature associated to raise the armrest on a chair. (5) demonstrates the reactions of the model to changes on such vector element.

By changing this element in the latent vector while keeping the other 31 constant, the user is able to produce a latent vector that can be decoded into a chair model without an armrest (5 left), or with different models of armrests (5 middle and right). However, we should assert that the behavior of this generative model is not completely predictable. The model is not guaranteed to generate exclusively acceptable chairs with usable armrests nor to generate novel armrests that are not present in the training dataset.

We trained this AE with point clouds generated based on geometric models from Shape Net data set[3]. Using AWS EC2 instances, we trained the AE on categories such as chairs, cars, hats, and tables. The trained AE model was deployed on the back-end of the DeepCloud application. To use it as a generative model, it receives a vector in $R^{32}$ and translates it to the respective point cloud in $R^{2048 \times 3}$, which represents 2048 points with related x, y, z coordination for each point.

6   ACADIA 2018                                                                                     DeepCloud  Bidgoli, Veloso

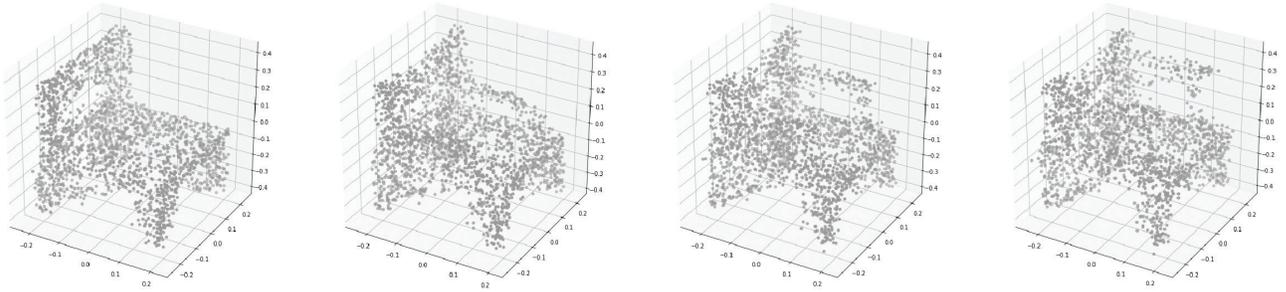

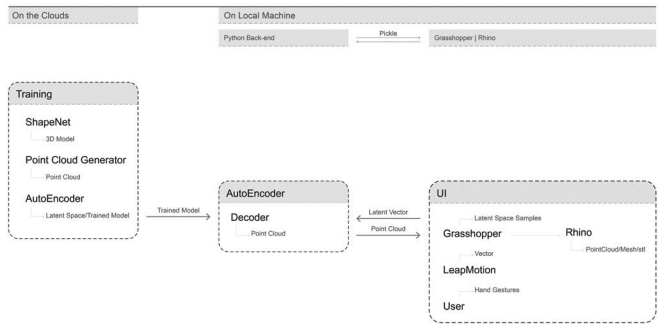

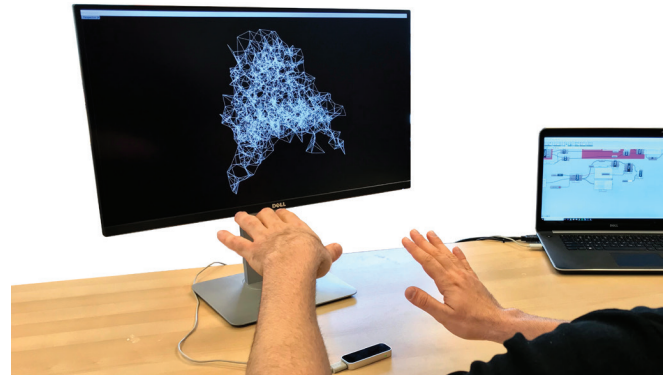

5    Effect of changing one element of the latent space vector on the decoded point cloud

6    System architecture of prototype 1

7    Interface of the prototype 1

Once the model was trained, we implemented a pipeline that feeds the forward pass of the decoder with data captured by a Leap Motion sensor integrated in the Grasshopper/Rhino modeling environment (6 and 7). To provide the user with more control over the interaction, we decided to only work with values that the user can control almost independently - i.e., palms' positions and angles and the total distance between fingers in each hand. The resulting 14-dimensional vector lets the user manipulate part of the 32-dimensional vector in the latent space. The vector is sent to the decoder to construct a point cloud, which is subsequently returned to the modeling environment.

After several tests, it was apparent that the bottlenecks in the Grasshopper-Python communication were significantly restricting the updating rate and decreasing the quality of the user experience. Despite the positive impact of generating complex models by hand gestures, the Leap Motion input added an extra level of complexity to the interaction. We observed that it was a cumbersome task maintaining a hand gesture while moving fingers to control other dimensions of the latent vector.

### Prototype 2

For the second prototype[4], we preserved the back-end from the previous experiment with minor improvements, but we opted to develop the prototype as a web-based application that could run on a standard modern web browser. It gave us the opportunity to design a platform-agnostic, fast, and scalable system for possible further developments. We updated the methods to improve the navigation in the latent space. Additionally, we substituted the Leap Motion with a MIDI mixer tool as a more intuitive user interface.

In the front-end (8 and 10), DeepCloud has a web-based interface that enables the user to manipulate the latent space representation and generate new point clouds. The user can navigate in the 3d space of the point cloud using the mouse, while the MIDI mixer controller with sliders and knobs provide an intuitive exploration of multiple transformations of the latent space.

### User Workflow

A. First, the users can choose between two modeling functionalities:

• Select an existing object from the database as a starting point for a new model and manipulate its features by modifying its latent vector.
• Select a group of objects and interpolate among them.

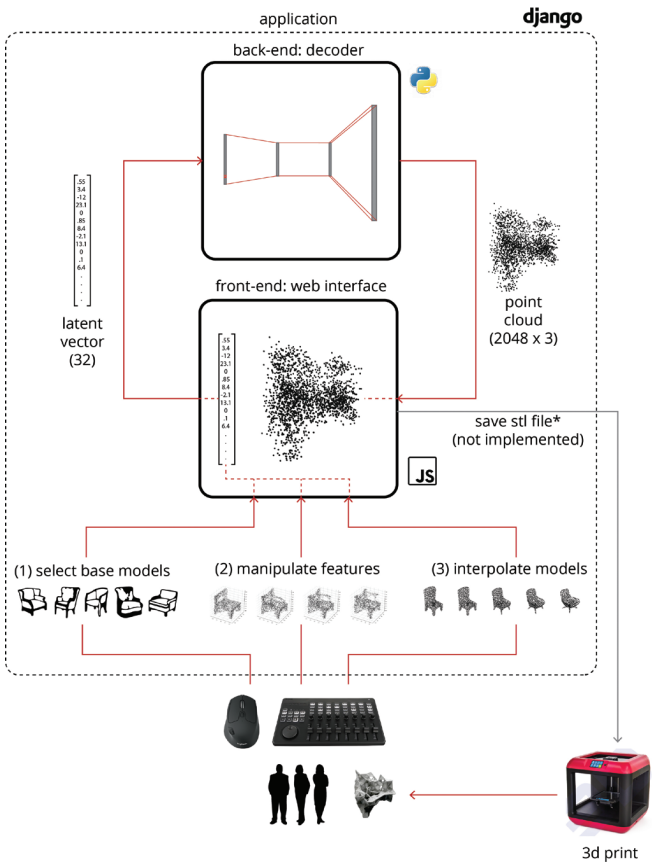

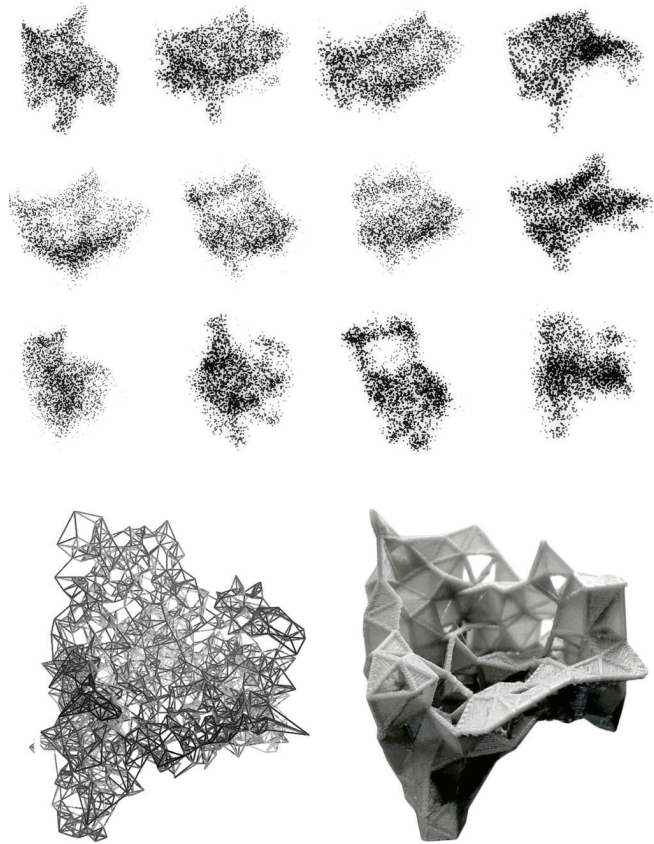

8   System architecture of prototype 2.

9   Top: chairs generated using DeepCloud; bottom left: structure; bottom right: 3d printed chair with lower resolution.

This enables the combination of multiple models to generate a hybrid with shared characteristics.

B. This will redirect user to the editing interface, containing editing tools and a 3d point cloud viewer (10 top).

C. In the case of the feature manipulation:

- The user will have access to eight sliders in the editing environment. Each of the sliders of the mixer is used to modify values to a vector $t$, representing a transformation in the latent space, while the adjacent knob could be used to fine-tune it. The range of the sliders is proportional to the interval of the corresponding latent values in the database.
- The resulting vector is added to the latent vector $f$ (representing the original model selected by the user), creating a new vector $x$ that represents the transformed model. This new model will be reflected in the point cloud viewer in real-time.
- In the prototype, the sliders and knobs only control the first eight values of the vector $t$, but potentially all its components can be edited. For most of the components of the feature space, these $t_i$ values represent an identifiable transformation of the model, which are depicted in the interface with animated GIFs (10 middle row). In our feature space for the class of cars, the component $t_5$ was associated with the addition of a spoiler to the trunk. For the chairs, $t_2$ was tied to increasing the size of the model and adding an opening to the back of the seat.

D. For the interpolation of $n$ objects:

- The user will have access to $n$ sliders. Each slider controls the weight of each model in the linear combination of their respective latent vectors.
- By manipulating each slider, the user can apply different weights to each object and control its influence in the final resulted model.
- Changes in the slider will provide a weight vector $w$ in $R^n$, which is normalized and multiplied by the matrix ($V$) that contains all the latent vectors of the selected models ($v_0, v_1, …, v_{n-1}$), resulting in the hybrid latent vector $h$.
- The resulting latent vectors for both operations ($x$ or $h$) are sent to the AE, which returns the corresponding point cloud on the web-interface in real-time (10 bottom row). (9) represents some examples of chair models

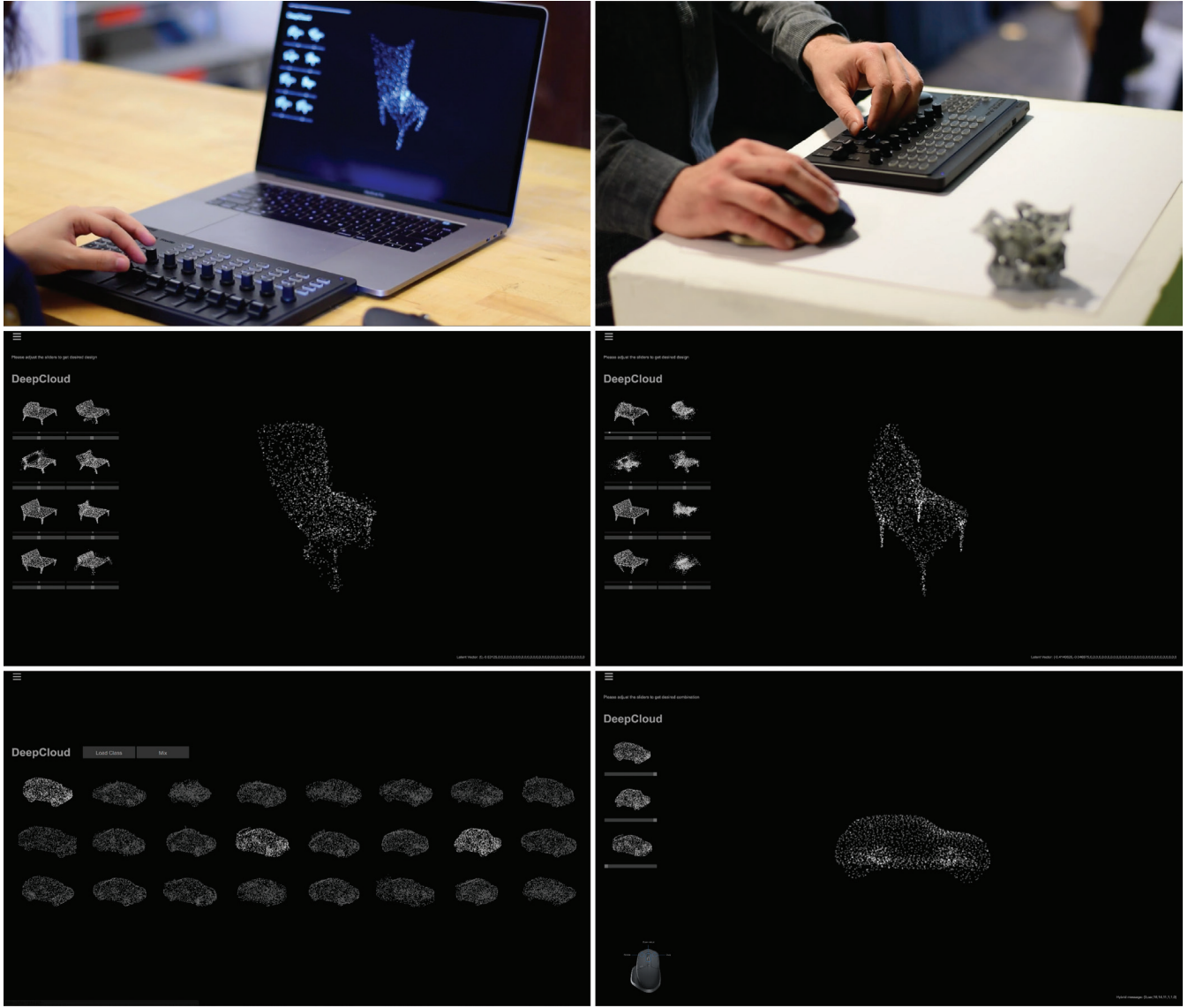

10  Interface of prototype 2. Top row: physical interface; Middle row: feature manipulation; Bottom row: hybridization/interpolation of models.

developed using DeepCloud

## FUTURE STEPS

The proposed system is an early prototype to study the affordances of generative models in design practice. For the next steps, the authors aim to address these aspects:

- One of the shortcomings of the proposed system is the lack of integration between the generated results and physical constraints of a given class. This could be improved by a physical optimization engine that satisfies specific constraints of a solution or flags impossible outputs.
- The other opportunity for improvement is the possibility of integrating semantic segmentation to the models. A similar approach has been demonstrated by Yumer and Kara for mesh geometries using statistical methods (2014). For example, in the case of chairs, the model can be trained to distinguish between legs, seat, back and handles, enabling independent modeling of the parts.
- Improve the system with a more user-friendly representation of final artifacts, such as meshes.

## DISCUSSION

This project is built and trained around a publicly available dataset of 3d models. The challenge of biased datasets is one of the primary sources of concern in the ML practice. Such databases are potentially weighing in favor of disproportionately dominant classes in their distribution.

In design, bias is a double-edged sword. On the one hand, selecting a limited and exclusive data set is an opportunity

TOPIC (ACADIA team will fill in)    RECALIBRATION  ON IMPRECISION AND INFIDELITY    9

for designers. They can curate their own design culture by narrowing down on a specific data-scape and exploring it in much deeper detail. On the other hand, instances that have been left out of the dataset will not emerge in the results. A sparse data-scape will probably result in a sparse design space that lacks diversity.

The dilemma of transparency and scope is common among current CAD users and computational designers. In a classic parametric modeling approach, parameters are explicitly associated with various design features via a graph. In such workflows, the user develops a solid grasp on different aspects of the parametric model and the interactions between parameters and features. This requires experienced users with a specific repository of skill sets to design, implement, maintain, and use parametric models. In contrast, ML-based models liberate the user/designer from orchestrating the parametric relations between model and the design features. Compared with the parametric modeling workflow, this one does not require highly-trained users in all parts of its life cycle. However, it is black box, so even a keen and experienced developer cannot understand how the model works. It is a trade-off of transparency in favor of end-user convenience.

Concerning the scope, in parametric modeling, the design space is restricted to the dependencies of the underlying graph, which is explicitly designed. Drastic topological or conditional variations of the model generally require the development of a new parametric model. ML-based models trained on diverse databases can automatically explore a more extensive variety of topologies that would be hard to model in a single parametric model.

Nonetheless, these conveniences come at a price. With current ML techniques, we trade the transparency of parametric design for a black box model. Parameters in the latent space are neither explicitly associated with any specific design features nor are they guaranteed to grasp a meaningful, unique feature. In such a scenario, the user should search for desired effects in a trial and error cycle and hold on to those effects in order to control the model. It is worth mentioning that each training session will shuffle the mapping of latent vectors, so they may not represent the same features.

## ACKNOWLEDGMENTS

The authors are grateful to Prof. Eunsu Kang, Prof. Barnabas Poczos for supervising this project in Art and Machine Learning course at Carnegie Mellon University, School of Computer Science. We would like to thank Shenghui Jia for his great contribution to the 3d printing process. We would like to express our gratitude to the CMU's Computational Design Lab (CodeLab) for its generous support and CNPq (National Council for Scientific and Technological Development), which partially supported this research.

## NOTES

1. For example, Pix2Pix (Isola et al. 2017) is a conditional GAN used for paired image-to-image translation. CycleGAN is a GAN that solves unpaired image-to-image translation (Zhu et al. 2017). Progressive Growing GAN (PGGAN) learns with a database of celebrities faces and can synthesize faces of celebrities (Karras et al. 2017). Deep Attention GAN (DA-GAN) uses a source image and a target pose to solve pose morphing (Ma et al. 2018).
2. The code of the project and demos of both prototypes are available on project's github page: https://github.com/Ardibid/DeepCloud
3. The point cloud data was originally generated by Achlioptas et. al and can be accessed from: https://www.dropbox.com/s/vmsdrae6x5xws1v/shape_net_core_uniform_samples_2048.zip. The model was trained using the code snippit provided by Achlioptas et al. on the project's github: https://github.com/optas/latent_3d_points

IMAGE CREDITS

All drawings and images by the authors.

**Ardavan Bidgoli** is a computational designer, robotics, and machine learning researcher at at the Comutational Design lab (CodeLab) at Carnegie Mellon University. He had been collaborating with Bentley Systems, Autodesk's OCTO team at Pier9 and Build Space reseach teams. He has a Bachelor of Architecture and a Master of Architecture from the University of Tehran, and a Master of Architecture in Design Computing from The Pennsylvania State University. Currently, he is a PhD student in computational design at Carnegie Mellon University, focusing on architectural robotics and machine learning.

**Pedro Veloso** Architect, educator, and researcher interested in design exploration methods supported by different modes of artificial intelligence, such as classical AI, bio-inspired AI, agent-based modeling and machine learning. He has a Bachelor of Architecture and Urbanism from the University of Brasilia. Master of Architectural Design from the University of Sao Paulo. Currently, he is a PhD candidate in computational design at Carnegie Mellon University, investigating the application of swarm intelligence in the design of spatial arrangements.